\documentclass[sigconf, screen]{acmart}
\usepackage{multirow}
\usepackage{makecell}
\AtBeginDocument{%
  }

\acmConference[]{}{}{}
\acmBooktitle{}
\renewcommand\footnotetextcopyrightpermission[1]{}
\begin{document}
\fancyhead{}

\title{M\textsuperscript{3}D-Stereo: A Multiple-Medium and Multiple-Degradation Dataset for Stereo Image Restoration}

\author{Deqing Yang}
\affiliation{%
  \institution{Shenzhen University}
  \city{Shenzhen}
  \country{China}}
\email{2410095078@mails.szu.edu.cn}

\author{Yingying Liu}
\affiliation{%
  \institution{Shenzhen University}
  \city{Shenzhen}
  \country{China}}
\email{liuyying@szu.edu.cn}

\author{Qicong Wang}
\affiliation{%
  \institution{Shenzhen University}
  \city{Shenzhen}
  \country{China}}
\email{2310295091@email.szu.edu.cn}

\author{Zhi Zeng}
\affiliation{%
  \institution{Chongqing Normal University}
  \city{Chongqing}
  \country{China}}
\email{zh406@cqnu.edu.cn}

\author{Dajiang Lu}
\authornote{Correspondence}
\affiliation{%
  \institution{Shenzhen University}
  \city{Shenzhen}
  \country{China}}
\email{ludajiang@szu.edu.cn}

\author{Yibin Tian}
\authornote{Correspondence}
\affiliation{%
  \institution{Shenzhen University}
  \city{Shenzhen}
  \country{China}}
\email{ybtian@szu.edu.cn}

\renewcommand{\shortauthors}{Yang et al.}

\begin{abstract}
Image restoration under adverse conditions, such as underwater, haze or fog, and low-light environments, remains a highly challenging problem due to complex physical degradations and severe information loss. Existing datasets are predominantly limited to a single degradation type or heavily rely on synthetic data without stereo consistency, inherently restricting their applicability in real-world scenarios. To address this, we introduce M\textsuperscript{3}D-Stereo, a stereo dataset with 7904 high-resolution image pairs for image restoration research acquired in multiple media with multiple controlled degradation levels. It encompasses four degradation scenarios: underwater scatter, haze/fog, underwater low-light, and haze low-light. Each scenario forms a subset, and is divided into six levels of progressive degradation, allowing fine-grained evaluations of restoration methods with increasing severity of degradation. Collected via a laboratory setup, the dataset provides aligned stereo image pairs along with their pixel-wise consistent clear ground truths. Two restoration tasks, single-level and and mixed-level degradation, were performed to verify its validity. M\textsuperscript{3}D-Stereo establishes a better controlled and more realistic benchmark to evaluate image restoration and stereo matching methods in complex degradation environments. It is made public under LGPLv3 license.

\end{abstract}

% <concept_id>10002951.10003227.10003251.10003256</concept_id>
% <concept_desc>Information systems~Multimedia content creation</concept_desc>
% \ccsdesc[500]{Information systems~Multimedia content creation}



\maketitle
\section{Introduction}
Image restoration in degraded environments, including underwater \cite{Li2021UnderwaterIE,10238432}, haze/fog~\cite{Ren2016SingleID,He2009SingleIH}, and low-light conditions~\cite{Wei2018DeepRD,Guo2020ZeroReferenceDC}, has become increasingly important for applications such as autonomous navigation, marine exploration, and virtual reality. Although monocular restoration has been widely studied for these degradation scenarios~\cite{He2009SingleIH,Wei2018DeepRD}, it ignores the geometric consistency available in stereo imaging and often fails to recover fine details under severe degradation. Stereo restoration offers a promising alternative that uses cross-view fusion and geometric constraints to compensate for information loss.

However, the development of stereo restoration is challenged by the lack of suitable benchmark datasets. Existing stereo datasets, such as KITTI~\cite{Geiger2013VisionMR}, mainly target disparity estimation and lack clear pixel-aligned images for photometric evaluation. In contrast, existing restoration datasets, such as UIEB~\cite{Li2019AnUI} and O-HAZE~\cite{Ancuti2018OHAZEAD}, are largely monocular and therefore unsuitable for studying stereo-consistent restoration. In addition, synthetic data cannot faithfully reproduce the complex physical effects of real degraded environments, such as multiple scatter and photon noise~\cite{Akkaynak2019SeaThruAM}. As a result, current datasets typically suffer from one or more limitations: (1) focusing on only a single degradation scenario; (2) relying predominantly on synthetic data; and (3) lacking fine-grained control over degradation severity.

To address these limitations, we introduce M\textsuperscript{3}D-Stereo (Multiple-Medium, Multiple-Degradation Stereo), a dataset for stereo image restoration. It covers four realistic degradation scenarios: underwater scatter (UWST), haze/fog scatter (HZST), underwater low-light (UWLL), and coupled haze and low-light (HZLL). Each scenario is divided into six progressive degradation levels (D1–D6), allowing controlled evaluation under increasing severity of degradation.

The dataset was built using a custom acquisition platform with two calibrated stereo camera systems and a turbidity-controllable imaging chamber. To enrich structural and semantic diversity, we constructed scenes using corals, rocks, aquatic plants, miniature vehicles, and figurines~\cite{Scharstein2014HighResolutionSD,wang2025itw}. For every pair of degraded stereo images, a clear reference was captured without degradation under the same scene and camera configuration~\cite{Li2019AnUI}. This ensures pixel-wise alignment for photometric evaluation and also enables accurate disparity ground truths (GTs) to be derived from clear stereo pairs, providing useful geometric supervision for future stereo restoration and stereo matching studies~\cite{Scharstein2001ATA,Kendall2017EndtoEndLO}. 

Compared with existing datasets, M\textsuperscript{3}D-Stereo offers several distinct advantages: (1) It provides aligned stereo image pairs under realistic degradations, allowing geometry-aware learning and evaluation. (2) It covers multiple media, including both underwater and haze/fog, within a unified benchmark. (3) It includes six controlled degradation levels for each scenario for fine-grained performance analysis. (4) It simultaneously provides photometric and geometric GTs without degradation.

We further evaluate two representative existing methods for stereo image restoration under various degradation conditions using the dataset~\cite{Wang2019Flickr1024AL} as benchmark. By providing aligned stereo pairs with controllable degradations, M\textsuperscript{3}D-Stereo supports research on geometry-aware restoration, where geometric consistency between stereo views serves as an additional constraint for recovering degraded images through cross-view information fusion~\cite{Wang2019LearningPA,Chu2022NAFSSRSI}. We expect that M\textsuperscript{3}D-Stereo will facilitate future research in stereo image restoration and also support more challenging tasks, such as color-depth joint restoration \cite{lu2025multi} and stereo matching~\cite{Wang2025RoSeRS}  in adverse environments.

\section{Related Work}
To put the proposed dataset in context, we classify existing relevant degradation datasets into four main categories: \textit{Monocular Synthetic}, \textit{Monocular Real}, \textit{Stereo Synthetic}, and \textit{Stereo Real}. A detailed summary of the datasets is given in Table~\ref{tab:1}.

\begin{table*}[t]
\centering
\caption{Comparison of imaging degradation datasets (UW: underwater; GT: ground truth). }
\label{tab:1}
%\resizebox{\textwidth}{!}
{
\begin{tabular}{l l l c c c c c l}
\toprule
Category & Dataset & Degradation & Resolution & Syn/Real & Size & Aligned & GT & Task \\
\midrule

\multirow{6}{*}{Synthetic}
& HazyKITTI2012~\cite{Wang2024ProgressiveSI,Geiger2013VisionMR}& Haze & 1242$\times$375 & Syn & 778 & Yes& Yes & Stereo Image Restoration \\
& HazyKITTI2015~\cite{Wang2024ProgressiveSI,Geiger2013VisionMR}& Haze & 1242$\times$375 & Syn & 800 & Yes& Yes & Stereo Image Restoration \\
& LLHolopix50~\cite{Zhao2024LowlightSI} & Low-light & 1280$\times$720 & Syn & 1189 & Yes& Yes & Stereo Image Restoration \\
& LLFlickr2014~\cite{Zhao2024LowlightSI,Thomee2015YFCC100M} & Low-light & 600$\times$1696 & Syn & 391 & Yes& Yes & Stereo Image Restoration \\
& LLKitti2015~\cite{Zhao2024LowlightSI,Menze2015ObjectSF} & Low-light & 1242$\times$375 & Syn & 400 & Yes& Yes & Stereo Image Restoration \\
& UWStereo~\cite{Lv2024UWStereoAL} & UW & 1280$\times$720 & Syn & 29568 & Yes& No & Stereo Matching \\
\midrule

\multirow{2}{*}{Real}
& SQUID~\cite{Berman2018UnderwaterSI} & UW scatter & 1827$\times$2737 & Real & 57 & Yes & No & Stereo Image Restoration \\
& DrivingStereo~\cite{Yang2019DrivingStereoAL} & Fog/Driving & 1762$\times$800 & Real & 500 & Yes& No & Stereo Matching \\
\midrule

\multirow{5}{*}{M\textsuperscript{3}D-Stereo}
& UWST subset (Ours) & UW scatter & 1920$\times$1080 & Real & 1536 & Yes & Yes & Stereo Image Restoration \\
& UWLL subset (Ours)& UW low-light & 1920$\times$1080 & Real & 1536 & Yes & Yes & Stereo Image Restoration \\
& HZST subset (Ours)& Haze/Fog & 1920$\times$1080 & Real & 2112 & Yes & Yes & Stereo Image Restoration \\
& HZLL subset (Ours)& Haze low-light & 1920$\times$1080 & Real & 2112 & Yes & Yes & Stereo Image Restoration \\
& \textbf{Combined (Ours)} & \textbf{4 scenarios} & \textbf{1920$\times$1080} & \textbf{Real} & \textbf{7296} & \textbf{Yes} & \textbf{Yes} & \textbf{Stereo Image Restoration} \\
\bottomrule
\end{tabular}
}
\end{table*}

\subsection{Synthetic Monocular Degradation Datasets}
Such datasets are widely used because they provide precise control over degradation parameters and, in some cases, auxiliary information such as depth. Representative examples, such as SynFog~\cite{Xie2024SynFogAP}, generate degraded images by applying physically inspired models or rendering techniques to clear images to simulate conditions such as low-light or scatter. Although these datasets offer clear advantages in scalability and controllability, they suffer from two fundamental limitations. First, they are restricted to monocular settings and do not provide stereo image pairs, making them unsuitable for geometry-aware learning~\cite{Wang2019LearningPA}. Second, synthetic rendering often fails to capture the complex physical processes of real-world environments, such as multiple scatter, wavelength-dependent attenuation, spatially varying illumination, and device noise~\cite{Shao2020DomainAF}. As a result, models trained on such datasets often generalize poorly to real scenes. In particular, the inability to faithfully reproduce the combined effects of scatter and absorption in real physical environments leads to a substantial domain gap.

\subsection{Real Monocular Degradation Datasets}
This category of datasets captures visual degradations in the real physical environments. For example, O-HAZE~\cite{Ancuti2018OHAZEAD} and Dense-Haze~\cite{Ancuti2019DenseHazeAB} use dedicated haze generation systems to create realistic atmospheric conditions, while datasets such as BeDDE~\cite{Zhao2020DehazingER}, RUIE~\cite{Liu2019RealWorldUE}, and UIEB~\cite{Li2019AnUI} collect natural images in foggy or underwater environments. Although these datasets offer high photometric fidelity, they are fundamentally limited by their monocular nature. They do not provide GTs and stereo correspondences, which restricts their use in geometry-aware tasks~\cite{Wang2025RobuSTereoRZ}. In addition, they often model degradation only as a binary condition, e.g., degraded versus clear, or as a coarse category, and therefore lack the fine-grained and controllable degradation levels needed for more systematic evaluation.

\subsection{Synthetic Stereo Degradation Datasets}
To evaluate stereo matching in adverse conditions while retaining perfectly dense matching GTs, the dominant strategy is to synthesize degradation effects on top of clear stereo image pairs. Representative examples include HazyKITTI2012~\cite{Wang2024ProgressiveSI,Geiger2013VisionMR}, LLHolopix50~\cite{Zhao2024LowlightSI}, LLFlickr2014~\cite{Zhao2024LowlightSI,Thomee2015YFCC100M}, and LLKitti2015~\cite{Zhao2024LowlightSI,Menze2015ObjectSF}, which simulate haze/fog or low-light degradations in driving scenes. More recently, UWStereo rendered a large-scale underwater stereo dataset using Unreal Engine~\cite{Lv2024UWStereoAL}. Although these datasets provide large-scale stereo pairs together with pixel-accurate disparity GTs, synthetic rendering still struggles to reproduce the device noise, non-uniform illumination, and complex medium effects present in real environments. As a result, models trained on such benchmarks often experience substantial performance degradation~\cite{Zhang2019DomaininvariantSM} when deployed in real adverse conditions.

\subsection{Real Stereo Degradation Datasets}
This category includes a small number of pioneering datasets that capture stereo image pairs in real-world environments. For example, SQUID~\cite{Berman2018UnderwaterSI} collected natural underwater stereo images, while DrivingStereo~\cite{Yang2019DrivingStereoAL} recorded driving scenes under various weather conditions. These datasets provide both physical realism and stereo observations. However, their main limitation lies in the uncontrollable nature of open-world environments, where photometric GTs are not available, and degradation factors such as fog density or water turbidity cannot be precisely adjusted. As a result, they do not provide systematically defined degradation levels, making fine-grained analysis of algorithm robustness under increasing degradation difficult~\cite{Bijelic2019SeeingTF}. In fact, achieving strict and progressive degradation levels for the same scene in natural weather is physically infeasible~\cite{Sakaridis2017SemanticFS}.

As discussed above, existing datasets are typically limited to a single degradation scenario and therefore do not reflect the complexity of the cross-domain encountered in real-world deployments~\cite{Jiang2024ASO}. Moreover, there exists a fundamental trade-off between physical realism and controllability~\cite{Sakaridis2017SemanticFS}: synthetic datasets provide precise control but lack realism, whereas real-world datasets capture authentic degradations but do not offer systematic variations.

In contrast, M\textsuperscript{3}D-Stereo is intended to provide both comprehensiveness and controllability. It fills an important gap in stereo benchmarks for adverse environments and helps bridge underwater and atmospheric vision research within a unified framework. By alleviating the conventional conflict between realism and control, M\textsuperscript{3}D-Stereo integrates multiple media while maintaining strictly controlled progressive degradation levels. In addition, it provides high-quality aligned stereo pairs, photometric GTs, and accurate dense disparity GTs, even under coupled degradation conditions~\cite{Mayer2015ALD}. By overcoming the limitations of existing datasets, it establishes a new benchmark for evaluating stereo restoration and geometry-aware methods under complex real-world degradations.

\section{Dataset Construction}
This section describes the construction of the M\textsuperscript{3}D-Stereo, including the experimental platform, scene design, degradation generation, and data acquisition pipeline. 

\subsection{Experimental Setup and Scene Construction}
\begin{figure*}[t]
  \centering
  \includegraphics[width=\linewidth]{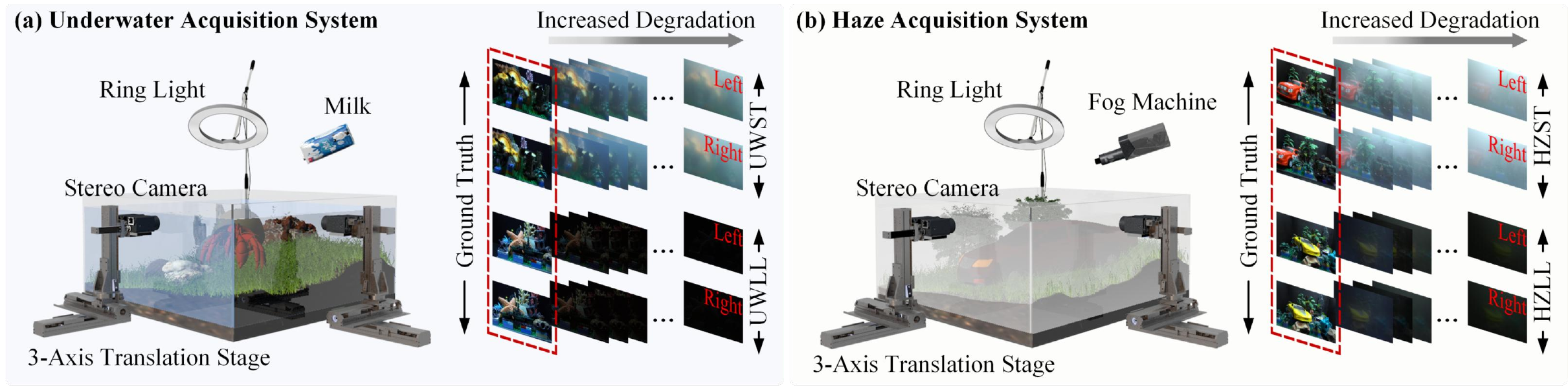}
  \caption{The M\textsuperscript{3}D-Stereo data acquisition platform. (a) Underwater stereo acquisition system: milk is added to the glass tank to simulate underwater scatter at varying concentrations, enabling capturing UWST and UWLL images. (b) Haze stereo acquisition system: a fog machine generates haze scenes of varying density within an enclosed space, enabling the acquisition of HZST and HZLL images. Degradation severity increases along the arrow direction.}
  \Description{Two data acquisition setups: (a) underwater system with a glass tank and milk-based scatter simulation, and (b) haze system with a fog machine in an enclosed space. Both use three-axis translation stages, ZED stereo cameras, and ring lights.}
  \label{fig:1}
\end{figure*}

Building on the experience of a previous small-scale study on underwater imaging \cite{wang2025itw, lu2025multi}, we redesigned the image acquisition platform to include a high-precision three-axis XYZ translation stage (XG100, Ruibo), two ZED stereo cameras (ZED Mini, Stereolabs), a ring light, and a custom glass tank of size $80 \times 80 \times 60$~cm$^3$. Figure~\ref{fig:1}(a) shows the underwater stereo acquisition System, which is used to capture UWST and UWLL images. Figure~\ref{fig:1}(b) presents the Haze stereo system for acquisition of HZST and HZLL images. The acquisition platform is placed in a closed room so that ambient light can be completely eliminated when room lights are off.

To ensure scene diversity and structural richness, we constructed multiple modular scenes underwater and in air. The underwater scenes contain rocks, corals, aquatic plants, artificial reefs, shipwreck models, and other small objects. These elements were arranged in different combinations to generate diverse geometric structures and occlusion patterns. In the air, we designed miniature scenes that contain vehicles, pedestrians, and trees, to simulate urban and natural environments under adverse weather conditions. Using modular scene components, the platform can be flexibly reconfigured, allowing the creation of a large number of scenes with different layouts and visual appearances. For each scene, before applying any degradation, we captured clear GT images using the same camera and pose to ensure pixel-level spatial alignment with the degraded images.

\subsection{Stereo Camera Calibration}
\begin{figure}[t]
\centering
\includegraphics[width=0.9\linewidth]{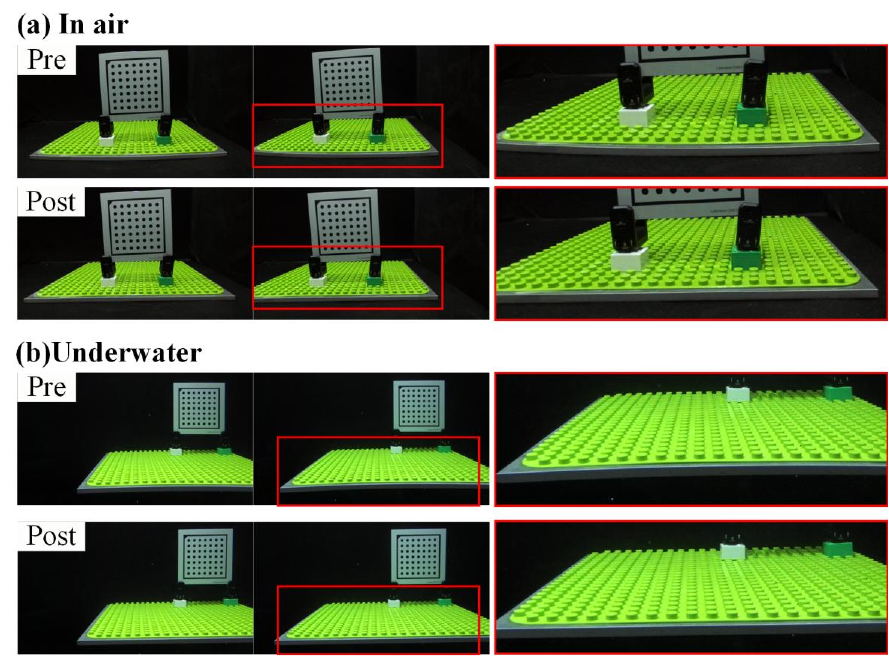}
\caption{Visualization of stereo calibration accuracy. (a) In air; (b) In clear water. Pre and Post denote the left and right image pairs before and after the calibration, respectively.}
\label{fig:2}
\end{figure}

Due to the significant refractive-index difference between air and water, conventional calibration parameters estimated in the air cannot satisfy the accuracy requirements of underwater data acquisition. To address this issue,  we performed a complete recalibration in both atmospheric and underwater environments following the strategy introduced by Li et al.~\cite{Li2023EvaluatingTE}. Specifically, we adopted Zhang's method~\cite{Zhang2000AFN} to calibrate the two ZED cameras in both clear water and air. Table \ref{tab:2} shows the reprojection errors (L\textsuperscript{rprj} and R\textsuperscript{rprj} for the left and right cameras, respectively) and the Y-offset (dY) after image rectification, the intrinsic and extrinsic parameters are provided in a separate file in the dataset. The calibration results are visually illustrated in Fig.~\ref{fig:2}.

\begin{table}[!h]
\centering
\caption{Stereo camera calibration accuracy.}
\label{tab:2}
\setlength{\tabcolsep}{3pt}
\begin{tabular}{ccccc}
  \toprule
  \textbf{Cam} & \textbf{Medium} & \textbf{L\textsuperscript{rprj} (px)} & \textbf{R\textsuperscript{rprj} (px)} & \textbf{dY (px)} \\
  \midrule
  \multirow{2}{*}{1}
  & Air      & 0.0337 & 0.0343 & 0.1604 \\
  & UW  & 0.0458 & 0.0427 & 0.1250 \\
  \midrule
  \multirow{2}{*}{2}
  & Air      & 0.0387 & 0.0382 & 0.3958 \\
  & UW  & 0.0471 & 0.0544 & 0.1358 \\
  \bottomrule
\end{tabular}
\end{table}

\subsection{Simulations of Physical Degradation}
A key feature of M\textsuperscript{3}D-Stereo is the rigorous control of physical degradation. We construct four distinct scenarios, each with six strictly controlled degradation levels from mild to severe (D1--D6).

\textbf{The UWST subset} was generated following the 3D TURBID~\cite{Duarte2016ADT, wang2025itw} protocol to simulate underwater scatter. The water turbidity was precisely controlled by progressively injecting a prepared milk solution into the tank. Specifically, 19 grams of milk powder (Xiyu Riji Skimmed Milk Powder) were dissolved in 1000 milliliters of water to form the turbid solution. This solution was introduced in six batches to create progressive degradation levels. An initial volume of 250 milliliters was injected for the first level, followed by five additional batches of 100 milliliters each. The cumulative injected volume at the highest degradation level, D6, reaches 750 milliliters. The UWST subset contains 256 image pairs for each degradation level.

\textbf{The UWLL subset} simulates underwater low-light degradation to mimic dark marine environments at different depths. Illumination was precisely controlled by a digital strobe controller (SJ-DPA60W24V, Shijue Factory) together with a ring light source (SJ-R18090-D80, Shijue Factory). The controller employs pulse-width modulation (PWM) dimming at a frequency of 86 kHz and supports 255 discrete brightness levels (0–255 in decimal), where lower values correspond to darker conditions. To ensure strictly progressive and physically grounded degradation, we fixed the PWM values at six levels, i.e., 11, 9, 7, 5, 3, and 1, and measured the corresponding illumination using a lux meter. For each fixed PWM setting, multiple readings were averaged to reduce measurement variability. The resulting illumination levels are 26.7, 20.8, 15.9, 10.5, 5.6, and 3.1 lux, respectively, providing a reproducible quantification of illumination intensity. The UWLL subset also contains 256 image pairs for each degradation level.

\textbf{The HZST subset} simulates haze/fog scatter degradation within a sealed physical space. A professional fogging system (FILMOG ACE portable fog machine, Ulanzi) with precisely controllable spray duration was used to generate haze under strictly progressive concentration levels. The initial spray duration was 10 s, and each subsequent level increases the duration by 5 s. As a result, the cumulative spray duration reaches 35 s at the highest degradation level (D6). By accurately controlling both the release dose and the diffusion time of the physical haze, this procedure creates a realistic fog scatter environment with approximately uniform distribution and well-defined progressive degradation. The HZST subset contains 352 image pairs for each degradation level.

\textbf{The HZLL subset} simulates the coupled degradation of haze/fog and low-light to reproduce highly challenging adverse conditions at night. Specifically, it combines selected levels from the two single-degradation settings. Haze levels D2, D4, and D6 were paired with low-light levels D1 and D3, resulting in six coupled degradation levels. Consequently, the composite levels D1--D6 correspond to: haze D2 + low-light D1, haze D2 + low-light D3, haze D4 + low-light D1, haze D4 + low-light D3, haze D6 + low-light D1 and haze D6 + low-light D3. This physically coupled design better reflects the nonlinear interaction between scatter and weak illumination and significantly increases the difficulty as well as the evaluation value of stereo image restoration. The HZLL subset also contains 352 image pairs for each degradation level.

To obtain high-quality GT stereo pairs that are pixel-wise aligned with observations across all degradation levels, we adopted a strictly controlled static-locking acquisition protocol. Both the stereo camera and all miniature objects in the scene were rigidly fixed to eliminate micro-motion. Clear stereo GT pairs were first captured under clean-medium and room illumination conditions, and these images serve as the reference for all subsequent acquisitions. The degradation media, such as milk or haze, were then gradually introduced, or the light level was progressively reduced. At each stable degradation level from D1 to D6, stereo image pairs were captured while the scene layout was kept unchanged. This protocol physically guarantees spatial consistency between the degraded observations and their corresponding GTs, thereby providing a reliable basis for stereo restoration and quantitative evaluation.

\begin{figure*}[t]
  \centering
  \includegraphics[width=0.95\linewidth]{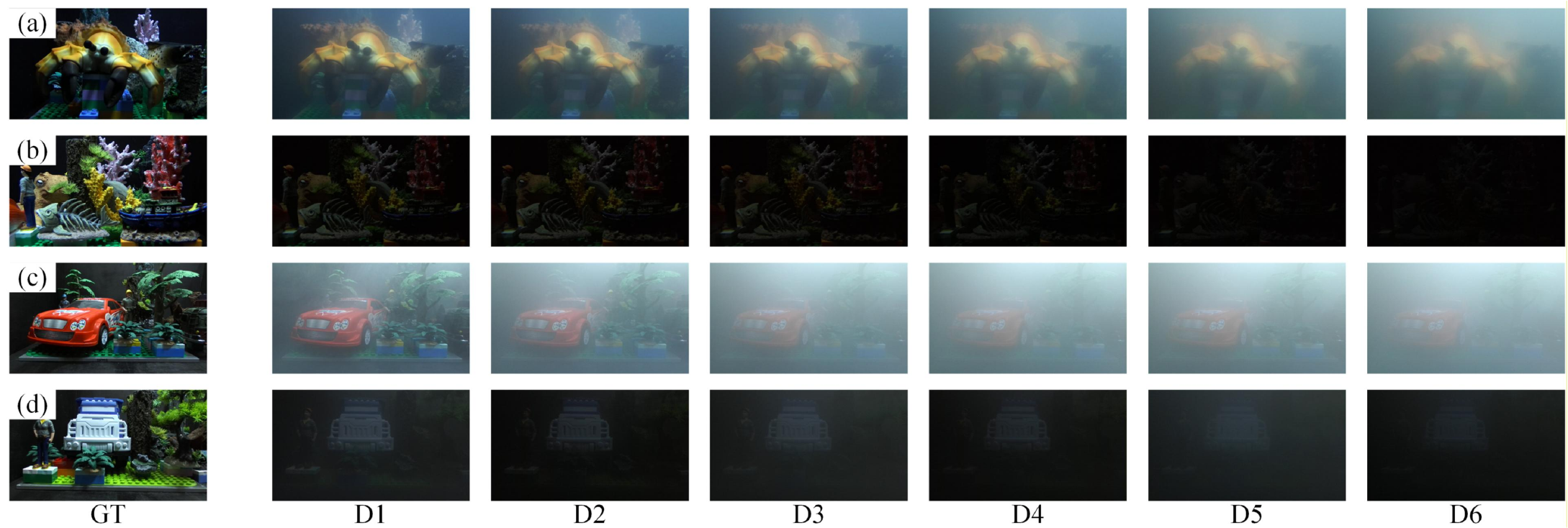}
  \caption{Sample images from the M\textsuperscript{3}D-Stereo dataset at degradation levels D1--D6. The leftmost column shows the clean GT. All displayed samples correspond to the left view only.(a) Underwater scatter (UWST). (b) Underwater low-light (UWLL). (c) Haze scatter (HZST). (d) Haze low-light (HZLL). Degradation severity increases from D1 to D6.}
  \Description{Four rows of sample images from the M2D-Stereo dataset showing progressive degradation from D1 to D6 across underwater scatter, underwater low-light, haze scatter, and haze low-light scenarios, with clean GT on the left. All images correspond to the left view.}
  \label{fig:3}
\end{figure*}

Table~\ref{tab:3} summarizes the degradation conditions and the number of image pairs in each subcategory. It should be noted that UWST and UWLL have the the same GTs of 256 pairs, and HZST and HZLL share the same GTs of 352 pairs. And Fig. ~\ref{fig:3} illustrates one example image for each of the degradation cases and its clear GT.

\begin{table}[!h]
\centering
\caption{Summary of M\textsuperscript{3}D-Stereo. Each level specifies the physical control parameter and the stereo image pairs.}
\label{tab:3}
\setlength{\tabcolsep}{1.8pt}
\begin{tabular}{ccccccccccc}
  \toprule
  \multirow{2}{*}{\textbf{Level}} & \multicolumn{2}{c}{\textbf{UWST}} & \multicolumn{2}{c}{\textbf{UWLL}} & \multicolumn{2}{c}{\textbf{HZST}} & \multicolumn{3}{c}{\textbf{HZLL}} \\
  \cmidrule(lr){2-3} \cmidrule(lr){4-5} \cmidrule(lr){6-7} \cmidrule(lr){8-10}
  & Milk(ml) & Pairs & Lux & Pairs & Fog(s) & Pairs & Fog(s) & Lux & Pairs \\
  \midrule
  GT & 0   & (256) & 173.7 & 256 & 0  & (352) & 0  & 141.3 & 352 \\
  D1 & 250 & 256 & 26.7 & 256 & 10 & 352 & 15 & 26.7 & 352 \\
  D2 & 350 & 256 & 20.8 & 256 & 15 & 352 & 15 & 15.9 & 352 \\
  D3 & 450 & 256 & 15.9 & 256 & 20 & 352 & 25 & 26.7 & 352 \\
  D4 & 550 & 256 & 10.5 & 256 & 25 & 352 & 25 & 15.9 & 352 \\
  D5 & 650 & 256 & 5.6  & 256 & 30 & 352 & 35 & 26.7 & 352 \\
  D6 & 750 & 256 & 3.1  & 256 & 35 & 352 & 35 & 15.9 & 352 \\
  \textbf{Total} & --- & \textbf{1536} & --- & \textbf{1792} & --- & \textbf{2112} & --- & --- & \textbf{2464} \\
  \bottomrule
\end{tabular}
\end{table}

\section{Experimental Evaluation}
Experimental validation consists of two tasks to evaluate the performance of image restoration methods on the dataset. Under a unified protocol, two representative stereo restoration methods, EPRRNet~\cite{Zhang2020BeyondMD} and PSIDNet~\cite{Wang2024ProgressiveSI}, were evaluated. Experiments were conducted on all four degradation scenarios: UWST, UWLL, HZST, and HZLL. All experiments used the dataset with a consistent training/testing split. To ensure a fair comparison, both methods were trained and tested in identical settings. Performance is evaluated using full-reference image-quality metrics, PSNR and SSIM~\cite{5596999}. 

\subsection{Single-Level Degradation}
\begin{figure*}[t]
  \centering
  \includegraphics[width=0.97\linewidth]{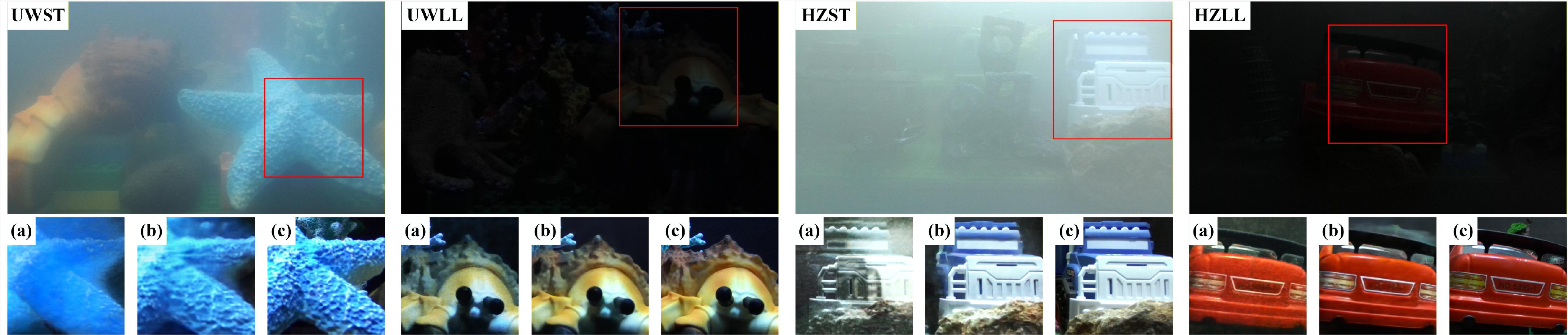}
  \caption{Restoration examples on the M\textsuperscript{3}D-Stereo dataset across four degradation scenarioss. From left to right: UWST, UWLL, HZST, and HZLL. The top row shows the full degraded images, and the bottom row shows zoomed-in comparisons of the red-box regions from: (a) EPRRNet restoration; (b) PSIDNet restoration; (c) clean GT.}
  \Description{Four-column comparison of restoration results across UWST, UWLL, HZST, and HZLL degradation types, with full degraded images on top and zoomed-in patches from EPRRNet, PSIDNet, and GT on the bottom.}
  \label{fig:4}
\end{figure*}

This task evaluates the restoration performance under different degradation levels. For each scenario, we selected three representative levels with clearly distinct intensities, namely D2, D4, and D6. The models were trained and tested independently on each level. The results are shown in Table~\ref{tab:4} and Fig.~\ref{fig:4}. A consistent trend is observed across all scenarios: as the degradation level increases from D2 to D6, both PSNR and SSIM decrease. This indicates that a stronger degradation causes more severe information loss and makes image restoration increasingly difficult. Similar trends have also been reported in previous studies on low-light enhancement~\cite{Guo2020ZeroReferenceDC,9609683} and dehazing~\cite{Li2019AnUI,Cai2016DehazeNetAE,He2009SingleIH} , where performance degradation is closely associated with reduced signal quality.

\begin{table}[!h]
\centering
\caption{Restoration results for single-Level degradation.}
\label{tab:4}
\setlength{\tabcolsep}{1.8pt}
\begin{tabular}{clcccccc}
\toprule
\multirow{2}{*}{\textbf{Scen}} & \multirow{2}{*}{\textbf{Model}} & \multicolumn{2}{c}{\textbf{D2}} & \multicolumn{2}{c}{\textbf{D4}} & \multicolumn{2}{c}{\textbf{D6}} \\
\cmidrule(lr){3-4} \cmidrule(lr){5-6} \cmidrule(lr){7-8}
 &  & PSNR$\uparrow$ & SSIM$\uparrow$ & PSNR$\uparrow$ & SSIM$\uparrow$ & PSNR$\uparrow$ & SSIM$\uparrow$ \\
\midrule
\multirow{2}{*}{UWST}
& EPRRNet & 18.46  & 0.6316 & 16.17  & 0.4989 & 13.92  & 0.4059 \\
& PSIDNet & \textbf{21.47}  & \textbf{0.7740} & \textbf{19.79}  & \textbf{0.6804} & \textbf{17.28}  & \textbf{0.5694} \\
\midrule
\multirow{2}{*}{UWLL}
& EPRRNet & 24.26  & 0.8266 & 23.02  & 0.8112 & 19.46  & 0.7076 \\
& PSIDNet & \textbf{25.06}  & \textbf{0.8614} & \textbf{25.08}  & \textbf{0.8499} & \textbf{23.20}  & \textbf{0.7835} \\
\midrule
\multirow{2}{*}{HZST}
& EPRRNet & 21.93  & 0.7408 & 17.30  & 0.5283 & 16.17  & 0.4428 \\
& PSIDNet & \textbf{24.81}  & \textbf{0.8381} & \textbf{21.29}  & \textbf{0.7307} & \textbf{19.71}  & \textbf{0.6556} \\
\midrule
\multirow{2}{*}{HZLL}
& EPRRNet & 18.80  & 0.5803 & 16.405 & 0.4310 & \textbf{15.795} & 0.4059 \\
& PSIDNet & \textbf{20.835} & \textbf{0.7101} & \textbf{18.015} & \textbf{0.5820} & 14.575 & \textbf{0.4977} \\
\bottomrule
\end{tabular}
\end{table}

Across all evaluated settings, PSIDNet consistently achieves higher PSNR and SSIM values than EPRRNet. The performance gap becomes more pronounced with increased severity of degradation. For example, under HZST at D4 and D6, PSIDNet maintains more stable structural similarity, indicating stronger robustness to moderate and severe scatter effects. A similar trend is observed for UWST, suggesting that PSIDNet preserves more image details under severe turbidity. These results show that stereo image restoration methods can maintain relatively stable performance at different degradation levels, while different network architectures exhibit noticeably different robustness under challenging conditions.

\begin{table}[t]
\centering
\caption{Image restoration results for mixed-level degradation. Six levels (D1--D6) were mixed for training and testing.}
\label{tab:5}
\setlength{\tabcolsep}{1.8pt}
\begin{tabular}{clcccccc}
  \toprule
  \multirow{2}{*}{\textbf{Scen}} & \multirow{2}{*}{\textbf{Model}} & \multicolumn{3}{c}{\textbf{Left view}} & \multicolumn{3}{c}{\textbf{Right view}} \\
  \cmidrule(lr){3-5} \cmidrule(lr){6-8}
  & & PSNR$\uparrow$ & SSIM$\uparrow$ & $\Delta E$$\downarrow$ & PSNR$\uparrow$ & SSIM$\uparrow$ & $\Delta E$$\downarrow$ \\
  \midrule
  \multirow{2}{*}{UWST}
  & EPRRNet & 18.60 & 0.6194 & 11.22 & 19.17 & 0.6314 & 11.03 \\
  & PSIDNet & \textbf{21.04} & \textbf{0.7348} & \textbf{8.20} & \textbf{21.07} & \textbf{0.7347} & \textbf{8.27} \\
  \midrule
  \multirow{2}{*}{UWLL}
  & EPRRNet & 20.61 & 0.6795 & 8.77 & 20.86 & 0.6790 & 8.50 \\
  & PSIDNet & \textbf{23.23} & \textbf{0.7795} & \textbf{6.36} & \textbf{23.63} & \textbf{0.7814} & \textbf{6.18} \\
  \midrule
  \multirow{2}{*}{HZST}
  & EPRRNet & 20.48 & 0.7380 & 9.33 & 20.88 & 0.7483 & 9.04 \\
  & PSIDNet & \textbf{24.37} & \textbf{0.8226} & \textbf{6.13} & \textbf{24.08} & \textbf{0.8223} & \textbf{6.52} \\
  \midrule
  \multirow{2}{*}{HZLL}
  & EPRRNet & 17.48 & 0.5066 & 13.05 & 17.78 & 0.5051 & 12.95 \\
  & PSIDNet & \textbf{18.48} & \textbf{0.6269} & \textbf{10.55} & \textbf{18.94} & \textbf{0.6261} & \textbf{10.28} \\
  \bottomrule
\end{tabular}
\end{table}

\subsection{Mixed-level Degradation}
This task evaluates a model's ability to handle mixed degradation levels using a single set of weights. For each subset, we combined all training samples from D1 to D6 to train one model. During testing, performance was evaluated separately at each degradation level and the final results are reported as average of all levels. In addition to PSNR and SSIM, we also use $\Delta E$~\cite{habekost2013which} to evaluate color fidelity, where a lower $\Delta E$ indicates better performance.

\subsection{Stereo Matching Evaluation}
\begin{figure}[!h]
  \centering
  \includegraphics[width=0.65\columnwidth]{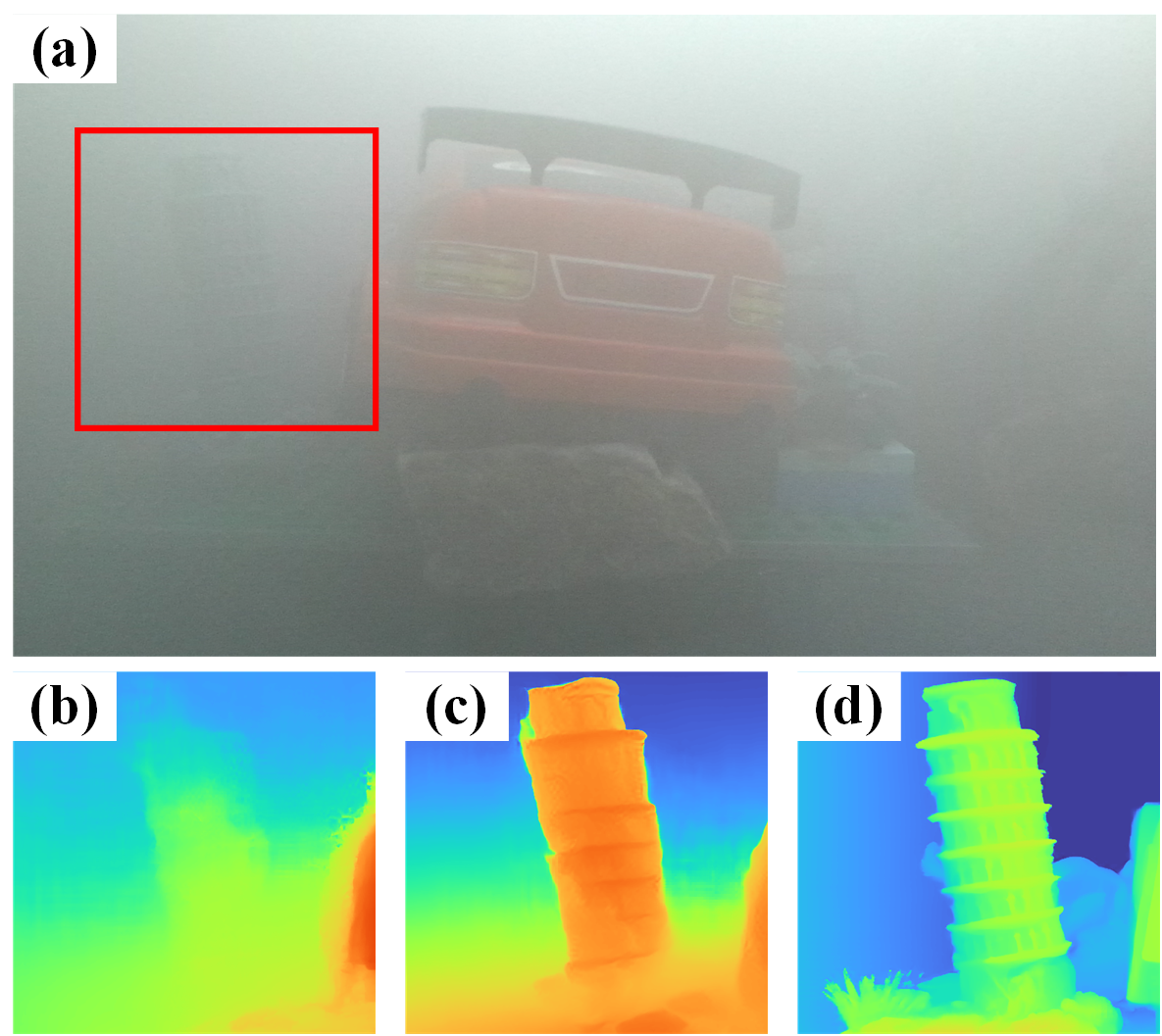}
  \caption{Impact of image restoration on stereo matching by pretrained FoundationStereo. (a) Degraded input (left view). (b) Depth map from degraded images. (c) Depth map from PSIDNet restored images. (d) Depth map from GTs.}
  \Description{Four-panel comparison showing a haze-degraded input and corresponding depth maps from stereo matching on the degraded image, PSIDNet restored image, and clean GT, demonstrating progressive improvement in depth estimation quality.}
  \label{fig:5}
\end{figure}

To show the benefit of stereo restoration for downstream stereo matching, we fed degraded images, PSIDNet restored results, and clean GTs into a pre-trained FoundationStereo model~\cite{Wen2025FoundationStereoZS} for depth estimation. As illustrated in Fig.~\ref{fig:5}, the depth map obtained by stereo matching on the degraded images exhibits severely distorted structures, with the background almost indistinguishable. In the depth map from restored images by PSIDNet, object contours become identifiable and depth layering is partially recovered. This comparison demonstrates that stereo image restoration can significantly improve the reliability of stereo matching under severe degradation.

\section{Access and Licensing}
The dataset will be made public on Huggingface~\url{https://huggingface.co/datasets/M3D-Stereo/M3D-Stereo}, under LGPLv3 license.

\section{Conclusion and Limitations}
This paper introduces M\textsuperscript{3}D-Stereo, a public dataset of multiple-medium stereo image restoration. It unifies underwater and atmospheric environments within a single framework and covers four degradation scenarios: UWST, UWLL, HZST and HZLL. Each scenario is further divided into six progressive degradation levels, enabling systematic evaluation under increasing degradation severity. It was acquired using a laboratory setup that balances physical realism with precise control over degradation factors. The dataset provides aligned stereo image pairs together with the corresponding GTs, enabling better evaluations of restoration performance. M\textsuperscript{3}D-Stereo was validated with two stereo restoration methods for single-level and mixed-level degradation settings as a benchmark. Results show that restoration performance consistently declines as degradation becomes stronger, while training on mixed degradation levels improves model robustness.

Despite the advantages, M\textsuperscript{3}D-Stereo has some limitations. First, it was acquired in a controlled environment with miniaturized objects and does not fully capture the scale of natural scenes. Second, although it covers multiple scenarios, its diversity remains limited compared to the real environment. Third, the current haze/fog and low-light coupled degradation is still limited in combinations. Future directions include extending the dataset to more complex environments, incorporating additional scenarios, such as rain and dust, and exploring its use in geometry-aware tasks such as stereo matching and color-depth joint restoration. We expect M\textsuperscript{3}D-Stereo to serve as a useful benchmark for research on stereo image restoration and stereo matching under complex degradations.

\begin{acks}
National Key R\&D Program of China (2024YFB4710600), LingChuang Research Project of China National Nuclear Corporation (CNNC-LCKY-2024-072), Shenzhen Science and Technology Innovation Commission (JCYJ20240813141402003) and Shenzen Talent Startup Funds (827-000954).
\end{acks}

\bibliographystyle{ACM-Reference-Format}
\bibliography{main}

\appendix

\end{document}